%% file: main.tex
\PassOptionsToPackage{numbers,square}{natbib}
\documentclass{article} 
\usepackage{nfam2025_workshop}
\setcitestyle{square,numbers,comma,sort&compress}
\usepackage{times}
\usepackage{booktabs}
\input{math_commands.tex}

\usepackage{adjustbox}
\usepackage{hyperref}
\usepackage{url}
\usepackage{float}
\usepackage{algorithm}
\usepackage{algorithmicx}
\usepackage{algpseudocode}
\usepackage{amsmath}
\usepackage{amssymb}
\usepackage{wrapfig}
\usepackage{multirow}

\usepackage{makecell}
\usepackage{wrapfig}
\newboolean{showcomments}
\setboolean{showcomments}{true}

\ifthenelse{\boolean{showcomments}}
  {\newcommand{\nb}[3]{
  {\color{#2}\small\fbox{\bfseries\sffamily\scriptsize#1}}
  {\color{#2}\sffamily\small$\triangleright~$\textit{\small #3}$~\triangleleft$}
  }
  }
  {\newcommand{\nb}[3]{}
  }

\title{Re:Frame -- Retrieving Experience From \\Associative Memory}

\author{%
    \textbf{Daniil Zelezetsky}$^{1}$\thanks{Equal contribution}
    \quad \textbf{Egor Cherepanov}$^{2,1*}$
    \quad \textbf{Alexey K. Kovalev}$^{2,1}$ 
    \quad \textbf{Aleksandr I. Panov}$^{2,1}$ \\
    $^1$MIPT, Dolgoprudny, Russia
    \quad $^2$AIRI, Moscow, Russia\\
    {\tt\small zelezetskii.dv@phystech.edu, \{cherepanov,kovalev,panov\}@airi.net}
}

\iclrfinalcopy 
\begin{document}


\maketitle

\input{sections/00_abstract}

\input{sections/01_introduction}
\input{sections/02_related_works}
\input{sections/03_background}
\input{sections/04_method}
\input{sections/05_results}
\input{sections/07_limitations}
\input{sections/08_conclusion}


\bibliography{nfam2025_workshop}
\bibliographystyle{unsrt}
\newpage
\appendix
\input{sections/appendix/appendix}

\end{document}

%% file: math_commands.tex
\usepackage{amsmath,amsfonts,bm}

\def\eqref#1{equation~\ref{#1}}

\def\1{\bm{1}}

\DeclareMathAlphabet{\mathsfit}{\encodingdefault}{\sfdefault}{m}{sl}
\SetMathAlphabet{\mathsfit}{bold}{\encodingdefault}{\sfdefault}{bx}{n}



%% file: sections/00_abstract.tex
\begin{abstract}
Offline reinforcement learning (RL) often deals with suboptimal data when collecting large expert datasets is unavailable or impractical. This limitation makes it difficult for agents to generalize and achieve high performance, as they must learn primarily from imperfect or inconsistent trajectories. A central challenge is therefore how to best leverage scarce expert demonstrations alongside abundant but lower-quality data. We demonstrate that incorporating even a tiny amount of expert experience can substantially improve RL agent performance.
We introduce \textbf{Re:Frame} (\textbf{R}etrieving \textbf{E}xperience \textbf{Fr}om \textbf{A}ssociative \textbf{Me}mory), a plug-in module that augments a standard offline RL policy (e.g., Decision Transformer) with a small external Associative Memory Buffer (AMB) populated by expert trajectories drawn from a separate dataset. During training on low-quality data, the policy learns to retrieve expert data from the Associative Memory Buffer (AMB) via content-based associations and integrate them into decision-making; the same AMB is queried at evaluation. This requires no environment interaction and no modifications to the backbone architecture. On D4RL MuJoCo tasks, using as few as 60 expert trajectories (0.1\% of a 6000-trajectory dataset), Re:Frame consistently improves over a strong Decision Transformer baseline in three of four settings, with gains up to +10.7 normalized points. These results show that Re:Frame offers a simple and data-efficient way to inject scarce expert knowledge and substantially improve offline RL from low-quality datasets.
\end{abstract}

%% file: sections/01_introduction.tex
\section{Introduction}
\input{figures/visual_abstract}
Memory is fundamental to human intelligence, enabling us to accumulate experiences, learn from interactions, and make informed decisions in complex environments~\citep{tulving2002episodic,SQUIRE2004171,BADDELEY2010R136}. It allows seamless integration of past knowledge with present observations to guide adaptive behavior~\citep{eichenbaum2017memory,parr2020prefrontal,parr2022active}. A central mechanism is associative memory—the ability to recall and apply relevant experiences by linking them to the current context~\citep{associative_mem,memory_humnan}, enabling efficient and robust decisions even with limited information.

In reinforcement learning (RL), significant advances have led to agents achieving superhuman performance in settings where exploration is possible and large amounts of high-quality data can be collected~\citep{Mnih2015HumanlevelCT,Silver2017MasteringTG}. However, in offline RL, agents must learn entirely from a fixed dataset without further environment interaction~\citep{levine2020offline}. This creates a fundamental limitation: the performance of offline RL algorithms depends directly on the quality of available data~\citep{janner2021offline,cherepanov2023recurrent,kachaev2025a}.  

In practice, collecting large datasets of expert demonstrations is rarely feasible. Safety, cost, and scalability constraints make it difficult to obtain sufficient high-return trajectories, especially in complex or real-world domains~\citep{kumar2020conservative}. Instead, practitioners typically face datasets dominated by medium- or low-quality behavior, with only a handful of expert trajectories available. This mismatch between abundant suboptimal and scarce expert data has motivated the development of methods that explicitly leverage mixed-quality datasets~\citep{nair2020awac, kostrikov2021offline, zhou2021plas}. Yet, effectively incorporating a very small number of expert trajectories into training remains a challenging open problem.

Motivated by the cognitive role of associative memory, we propose \textbf{Re:Frame} (\textbf{R}etrieving \textbf{E}xperience \textbf{Fr}om \textbf{A}ssociative \textbf{Me}mory), a simple and general framework for offline RL. Re:Frame augments an existing policy architecture (e.g., Decision Transformer (DT)~\citep{chen2021decision}) with an external \textbf{Associative Memory Buffer} (\textbf{AMB}) that stores only a tiny fraction of expert demonstrations (e.g., 0.1\% of the training dataset size). During learning on suboptimal data, the agent learns to retrieve and integrate relevant expert experiences through content-based associations, effectively using memory to compensate for the lack of exploration and the scarcity of expert trajectories. On standard D4RL~\citep{fu2020d4rl} benchmarks, this approach consistently improves over DT in the low-expert regime, with gains up to +10.7 normalized points.

Our contribution can be summarized as follows:
\begin{itemize}
    \item We introduce \textbf{Re:Frame}, an associative memory framework for offline RL that enables efficient retrieval of expert demonstrations, addressing the challenge of training with predominantly suboptimal datasets and only a handful of expert trajectories.  
    \item We demonstrate that Re:Frame can be seamlessly integrated with existing architectures, significantly improving decision-making performance in offline RL benchmarks while using as little as 0.1\% of expert data.  
\end{itemize}

%% file: figures/visual_abstract.tex
\begin{wrapfigure}{r}{0.5\textwidth}
    \vspace{-30pt}
    \centering
    \includegraphics[width=0.5\textwidth]{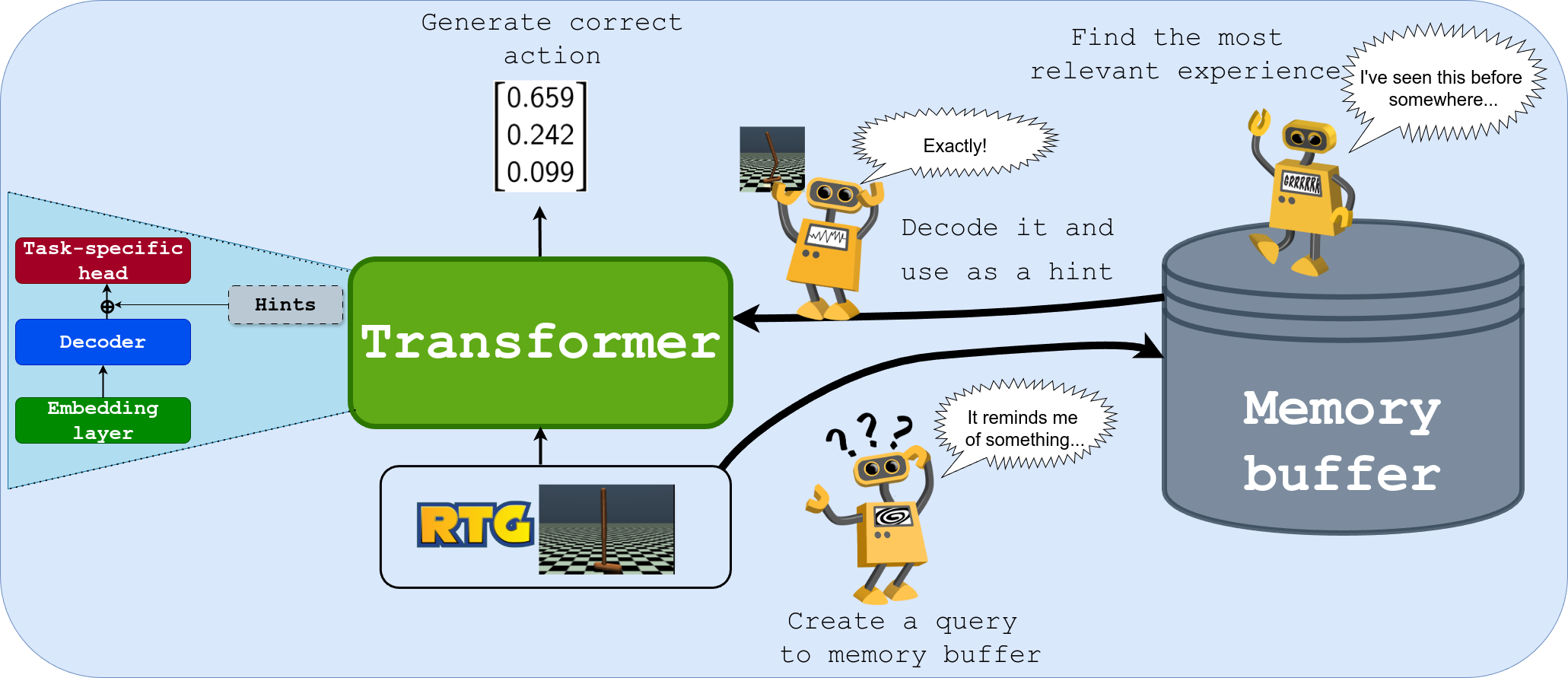}
    \vspace{-20pt}
    \caption{\textbf{Re:Frame overview.} A standard offline RL agent trained on suboptimal data is augmented with an \textit{Associative Memory Buffer} holding a few expert data. At each step, the agent queries memory, retrieves and decodes an expert data, and integrates it to guide the next action, enriching low-quality training with scarce expert knowledge.}

    \label{fig:Global_scheme}
    \vspace{-15pt}
\end{wrapfigure}

%% file: sections/02_related_works.tex
\section{Related Works}
\subsection{Associative Memory}
Associative memory has long inspired research in machine learning. Hopfield Networks~\citep{hopfield1982neural} pioneered associative recall, Neural Turing Machines~\citep{graves2014neural} added external memory with content- and location-based addressing, and Associative Recurrent Memory Transformer~\citep{rodkin2024associative} extended transformers with associative memory for long-term dependencies. The principles of associative memory have also been applied to RL~\citep{cherepanov2024unraveling}. Associative Memory Prioritized Experience Replay~\citep{li2022associative} speeds up prioritized replay, while Fast Weight Memory~\citep{schlag2020learning} augments Long Short-Term Memory~\citep{lstm} for meta-learning and associative inference. Self-attentive Associative Memory~\citep{le2020self} combines item and relational memory, and Associative Search Network~\citep{barto1981associative} enables autonomous sensory-motor learning from reinforcement signals. 
Episodic Reinforcement Learning with Associative Memory~\citep{Zhu2020EpisodicRL} builds a graph-based memory to connect experiences and propagate values. While prior work shows the promise of associative memory in RL, such methods are tied to specific architectures. In contrast, Re:Frame is architecture-agnostic, using a memory buffer that retrieves experiences by similarity rather than temporal order or graph links, enabling flexible, context-aware long-term recall.

\subsection{Offline RL with Limited Expert Data}
In offline RL, agents learn from a fixed dataset without exploration, so performance hinges on the quality and composition of trajectories. Classical methods - Advantage-Weighted Actor-Critic~\citep{nair2020awac}, Implicit Q-Learning (IQL)~\citep{kostrikov2021offline}, and Twin Delayed DDPG + Behavior Cloning (TD3+BC)~\citep{fujimoto2021minimalist} - constrain policy improvement toward the behavior distribution or weight updates by advantages/values, enabling learning from mixed-quality data with only limited expert demonstrations.

Recent work addresses the setting of few expert trajectories among many suboptimal ones. Guided Offline Reinforcement Learning~\citep{yang2023hundreds} adapts constraints using a small expert set, Hybrid Policy Optimization~\citep{yang2023hybrid} refines imperfect demos with online updates, and Optimal Transport for Offline Imitation Learning~\citep{luo2023optimal} aligns datasets with handfuls of expert examples. Equivariant Offline Reinforcement Learning~\citep{tangri2024equivariant} adds symmetry priors, Guided Data Augmentation~\citep{corrado2023guided} generates expert-like rollouts from small seeds. Re:Frame instead equips agents with a tiny expert associative memory (about 0.1\% of the dataset) that they learn to query, directly addressing the ``few experts + many suboptimal'' regime without modifying the backbone.


%% file: sections/03_background.tex
\input{algorithms/model}
\section{Background} 
\paragraph{Offline Reinforcement Learning.}
Offline RL learns policies from a fixed dataset $\mathcal{D}$ without further interaction with the environment, making it especially useful in domains where online data collection is expensive, unsafe, or impractical (e.g., robotics or healthcare). Each trajectory $\tau \in \mathcal{D}$ is a sequence of triplets $(r_t, o_t, a_t)$, containing the reward $r_t$, observation $o_t$, and action $a_t$ at each timestep. Decision Transformer (DT) reformulates RL as a sequence modeling problem by introducing the return-to-go $R_t = \sum_{k=t}^T r_k$, which represents the cumulative future rewards from timestep $t$. DT encodes $(R_t, o_t, a_t)$ as a sequence of tokens and feeds them into a transformer, which autoregressively predicts the next action conditioned on the desired return, the current observation, and the past trajectory. This formulation unifies RL with sequence prediction, enabling a single trained model to generate behaviors of varying quality simply by adjusting the return condition at inference time. However, a central limitation arises: when the dataset is dominated by suboptimal trajectories, DT struggles to recover expert-level performance, since it lacks direct access to sufficient high-quality demonstrations. In this work, we adopt DT as our base architecture because it is widely used in offline RL and makes this challenge explicit. Re:Frame addresses the limitation by augmenting DT with a compact expert memory buffer, allowing the model to incorporate and retrieve knowledge from even a very small set of expert trajectories while still training primarily on lower-quality data.

\paragraph{Autoencoder.} 
An Autoencoder (AE)~\citep{Rumelhart1986LearningIR} is a neural network that learns compact data representations through an encoder-decoder architecture. The encoder $f_\theta: \mathcal{X} \rightarrow \mathcal{Z}$ maps input data to a lower-dimensional latent space, while the decoder $g_\phi: \mathcal{Z} \rightarrow \mathcal{X}$ reconstructs the original input. Training minimizes the reconstruction loss $\mathcal{L}(\theta, \phi) = \|x - g_\phi(f_\theta(x))\|^2$. We leverage AE to create efficient encodings of agent experiences to support decision-making.

%% file: algorithms/model.tex
\begin{wrapfigure}[15]{r}{0.5\textwidth}
\vspace{-50pt}
\begin{minipage}{\linewidth}
\begin{algorithm}[H]
    \caption{Re:Frame-DT Integration}
    \label{alg:reframe-dt}
    \begin{algorithmic}[1]
    \Require AMB $\mathcal{B} \in \mathbb{R}^{T\times N}$, trajectory $\tau$
    \Ensure Predicted actions $\hat{a}$
    \State \textbf{Memory Retrieval:}
    \For{$(R_t,o_t,a_t)\in\tau$}
        \State $R'_t \gets \text{RtgEnc1}(R_t)$, $o'_t \gets \text{ObsEnc1}(o_t)$
        \State $h^*_t \gets \text{Linear}(\text{concat}(R'_t,o'_t))$
        \State $h'_t \gets \arg\min_{h\in\mathcal{B}} \|h^*_t-h\|_2^2$
        \State $a'_t \gets \text{ActDec1}(h'_t)$
        \State $a''_t \gets \text{Linear}(a'_t) \in \mathbb{R}^{1\times D}$
    \EndFor
    \State \textbf{Action Generation:}
    \State $\tau' \gets \text{Embed Sequence}(\tau)$
    \State $a^* \gets \text{Transformer}(\tau') \in \mathbb{R}^{T\times D}$
    \State $\hat{a} \gets \text{ActHead}(a^* + a'') \in \mathbb{R}^{T\times D}$
    \State \Return $\hat{a}$ with loss $\mathcal{L}(a,\hat{a})$
    \end{algorithmic}
\end{algorithm}
\end{minipage}
\end{wrapfigure}

%% file: sections/04_method.tex
\input{figures/model}
\section{Re:Frame Method}
\label{sec:reframe-method}
The proposed Re:Frame method employs a two-stage training strategy. In the first stage, we train an AE to construct the Associative Memory Buffer (AMB), a compressed repository of expert experiences (\autoref{fig:buffer}). Once trained, the AE’s parameters are frozen to ensure stable and consistent memory representations during subsequent learning. In the second stage, we build a decision-making framework that leverages the stored information in the AMB to enhance the agent’s performance (\autoref{fig:model}). This separation of memory construction and policy learning ensures that expert knowledge is distilled into a reliable latent structure before being integrated with policy optimization.

\input{tables/results}
\paragraph{Associative Memory Buffer.}  
Our method employs an Associative Memory Buffer (AMB) that compresses and stores a small set of expert demonstrations, enabling retrieval of high-quality experiences during both training and evaluation (see~\autoref{fig:buffer}). The key idea is to distill scarce expert trajectories into a compact and stable latent representation that the agent can later query.  

The buffer construction process begins by sampling the triplet $(R_t,o_t,a_t)$ at timestep $t$ from a dataset $\mathcal{D}$. Each component is encoded through a dedicated encoder, producing embeddings $(R'_t,o'_t,a'_t)$. These embeddings capture complementary aspects of agent-environment interaction: $R'_t$ encodes the cumulative reward signal, $o'_t$ captures the perceptual state, and $a'_t$ reflects the executed control action. The three embeddings are concatenated into a unified hidden state $h_t = \text{concat}(R'_t, o'_t, a'_t)$, which jointly represents reward, observation, and action.

To ensure compactness and robustness, $h_t$ is mapped through a linear projection into a lower-dimensional latent vector stored in the AMB $\mathcal{B}$. This projection acts as an information bottleneck, preserving task-relevant features while reducing redundancy. For stability, the latent code is decoded back into reconstructed components $(\hat{R}_t,\hat{o}_t,\hat{a}_t)$ using three independent decoders, each trained to minimize its own reconstruction loss:
$
\mathcal{L}_R = \|R_t - \hat{R}_t\|^2_2,
\mathcal{L}_o = \|o_t - \hat{o}_t\|^2_2,
\mathcal{L}_a = \|a_t - \hat{a}_t\|^2_2.
$

To reduce interference across modalities, each decoder is optimized with a separate optimizer. This modular setup ensures that returns, observations, and actions are faithfully preserved in the latent space, leading to stable memory encodings.  

By storing only a handful of expert trajectories - as little as $0.1\%$ of the dataset size - the AMB provides a lightweight yet highly informative memory. Through nearest-neighbor retrieval in the latent space, the agent can access relevant expert guidance during both training and evaluation. The availability of expert knowledge allows the agent to counterbalance the dominance of suboptimal data, improving robustness compared to approaches that merely fine-tune on scarce demonstrations.

\paragraph{Decision-making with Re:Frame.} 
To assess Re:Frame’s ability to exploit scarce expert data, we integrate it with the DT. Unlike vanilla DT, which relies only on the offline dataset, Re:Frame-DT augments decision-making with guidance from the AMB, a compact store of fewer than $0.1\%$ expert trajectories. The AMB can be queried during both training and evaluation, allowing the agent to counterbalance suboptimal data and obtain expert-informed corrections without large-scale demonstrations. As shown in~\autoref{fig:model} and Algorithm~\ref{alg:reframe-dt}, integration proceeds in two stages: \textit{Memory Retrieval}, where latent expert vectors are matched to the current context, and \textit{Action Generation}, where these signals are fused with DT’s predictions to refine the final action.

\paragraph{Memory Retrieval.} 
At each timestep $t$ of trajectory $\tau$, the agent performs a retrieval procedure to inject expert knowledge into decision-making. First, the returns-to-go $R_t$ and observations $o_t$ are encoded by the pre-trained AE into latent embeddings $R'_t$ and $o'_t$. These two components are concatenated into a joint query vector $\tilde{h}_t = \text{concat}(R'_t, o'_t)$, which represents the current context of the agent in latent space. 
Next, the query $\tilde{h}_t$ is projected through a linear transformation into the AMB latent space, yielding $h^*_t = W \tilde{h}_t + b$, where $W$ and $b$ are the learned projection parameters. This projection ensures that queries are aligned with the latent codes stored in the AMB. The AMB $\mathcal{B}$ contains a small set of expert trajectories stored as compact latent vectors $h$. To introduce expert guidance, we retrieve the most relevant expert memory $h'_t$ by performing nearest-neighbor search: $h'_t = \arg\min_{h \in \mathcal{B}} \| h^*_t - h \|_2^2.$ This retrieval corresponds to selecting the expert state that is most similar to the current query in latent space.  

Once the nearest memory $h'_t$ is obtained, it is decoded into a candidate expert action $a'_t = \text{Decoder}(h'_t)$, which represents the action suggested by the closest expert experience. To better integrate this retrieved action with the ongoing policy, a lightweight linear transformation refines the candidate into a correction vector: $a''_t = W_a a'_t + b_a$, where $W_a$ and $b_a$ are trainable parameters. This correction encodes the expert-informed adjustment that can be added to the agent’s predicted action, ensuring that retrieval contributes useful and context-aware guidance rather than directly overriding the policy.

\paragraph{Action Generation.} 
In parallel to memory retrieval, the input trajectory $\tau = \{(R_t, o_t, a_t)\}_{t=1}^T$ is processed through the standard DT encoders. This produces an embedded sequence $\tau' = \text{EmbedSequence}(\tau)$, which serves as the transformer’s input. The transformer then applies causal self-attention over $\tau'$ and maps it into action embeddings $a^* = \text{Transformer}(\tau')$, where $a^*_t$ denotes the dataset-driven action prediction at timestep $t$.  

To incorporate expert knowledge, the expert-informed correction vector $a''_t$, obtained during memory retrieval, is fused with the transformer’s prediction. Specifically, the two representations are combined additively: $\tilde{a}_t = a^*_t + a''_t$. This step ensures that the model’s action representation reflects both dataset-derived patterns and retrieved expert adjustments.  

Finally, the Action Head maps the fused embedding $\tilde{a}_t$ into the final action output: $\hat{a}_t = \text{ActionHead}(\tilde{a}_t)$. This design ensures that expert information refines but does not dominate the decision process, effectively blending dataset-driven learning with associative memory guidance from $\mathcal{B}$. As a result, the agent can leverage a very small number of expert demonstrations to consistently improve performance, without requiring large-scale expert datasets.

%% file: figures/model.tex
\begin{figure*}[t!]
\centering
\vspace{-20pt}
\begin{minipage}[t]{0.49\textwidth}
    \centering
    \includegraphics[width=\textwidth]{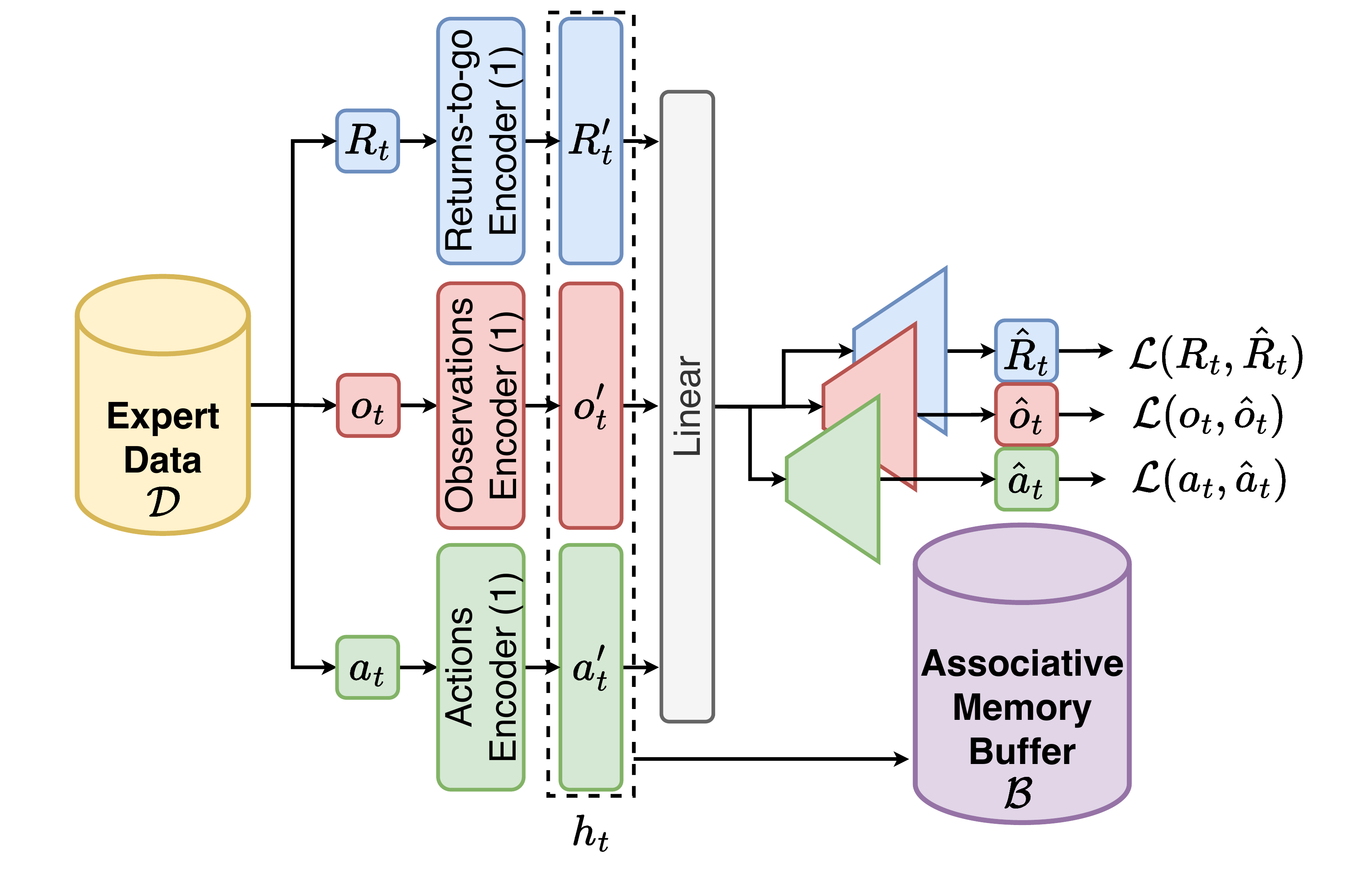}
    \vspace{-20pt}
    \caption{\textbf{Construction of the Associative Memory Buffer (AMB).} 
    Expert trajectories $(R_t, o_t, a_t)$ are encoded by an autoencoder into latent vectors $(R'_t, o'_t, a'_t)$, and concatenated into a hidden state $h_t$. 
    A linear projection maps $h_t$ into a compact latent representation, which is then decoded back to reconstruct $(\hat{R}_t, \hat{o}_t, \hat{a}_t)$ with component-wise reconstruction losses. 
    This training yields stable, information-preserving representations stored in the AMB as compact expert knowledge.
    During learning, the AMB retrieves expert data, letting the agent augment scarce high-quality data with abundant suboptimal trajectories.}

    \label{fig:buffer}
\end{minipage}%
\hfill
\begin{minipage}[t]{0.49\textwidth}
    \centering
    \includegraphics[width=\textwidth]{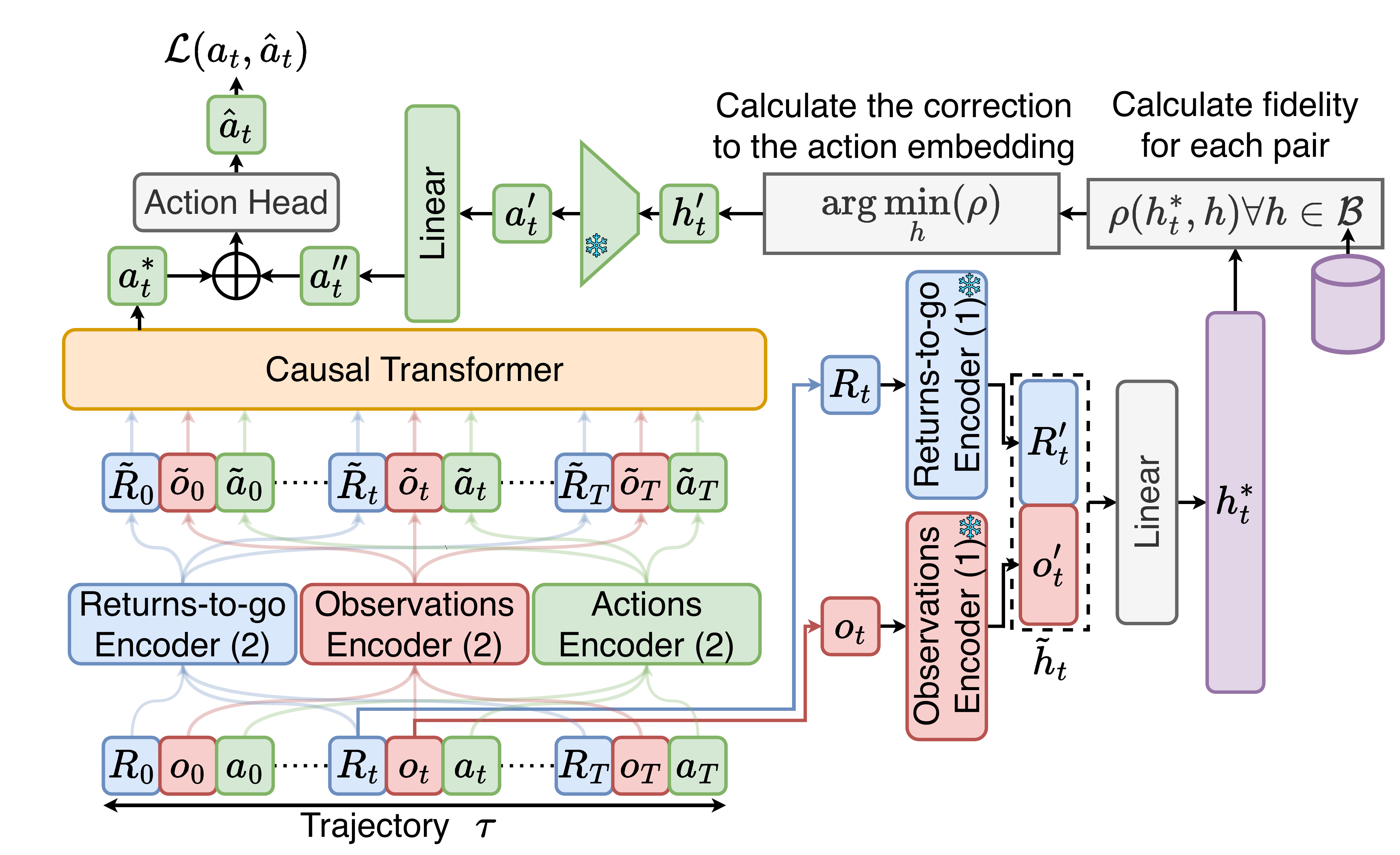}
    \vspace{-20pt}
    \caption{\textbf{Integration of Re:Frame with Decision Transformer (DT).} 
    At each timestep, the agent encodes current return-to-go $R_t$ and observation $o_t$ into a query vector $h^*_t$, which is projected into the AMB latent space. 
    The AMB is searched for the most relevant expert embedding $h$, minimizing the fidelity metric $\rho(h^*_t, h)$. 
    The retrieved memory is decoded into an expert-informed action candidate $a'_t$, refined via a correction vector $a''_t$, and combined with the DT action embedding $a^*_t$ through the Action Head to generate the final action $\hat{a}_t$. 
    This integration lets the transformer combine suboptimal training with expert guidance from a small memory.}
    \label{fig:model}
\end{minipage}
\vspace{-10pt}
\end{figure*}

%% file: tables/results.tex

\begin{table}[t]
\centering
\caption{\textbf{D4RL normalized scores (mean $\pm$ std).} AMB size is fixed to 60 trajectories (0.1\% of the dataset). 
\emph{Fine-tuned DT} denotes a DT fine-tuned directly on the same 60 expert trajectories used in Re:Frame's AMB. 
\emph{Swap expert @ eval} means the model is trained with a dataset-derived AMB but evaluated with an expert AMB. 
\emph{Expert train+eval} uses an expert AMB throughout. 
\emph{Dataset AMB} uses a dataset-matched (non-expert) AMB during both training and evaluation. 
Best per row in \textbf{bold}.}
\vspace{5pt}
\label{tab:main-results}
\resizebox{\textwidth}{!}{
\begin{tabular}{llccccc}
\toprule
\makecell[c]{Task} & 
\makecell[c]{Dataset} & 
\makecell[c]{DT} &
\makecell[c]{Fine-tuned\\ DT} &
\makecell[c]{Re:Frame:\\ swap expert @ eval} &
\makecell[c]{\textbf{Re:Frame:}\\ \textbf{expert train+eval}} &
\makecell[c]{Re:Frame:\\ dataset AMB} \\
\midrule
Hopper   & MR & 82.7 $\pm$ 7.0 & \textbf{84.8 $\pm$ 2.2} & 33.9 $\pm$ 1.6 & 69.7 $\pm$ 1.7 & 41.8 $\pm$ 9.0 \\
Hopper   & M  & 67.6 $\pm$ 1.0          & 68.3 $\pm$ 1.2 & 67.2 $\pm$ 2.3 & \textbf{78.3 $\pm$ 1.6} & 67.9 $\pm$ 4.9 \\
Walker2d & MR & 66.6 $\pm$ 3.0          & 65.6 $\pm$ 2.1 & 64.4 $\pm$ 7.7 & \textbf{69.8 $\pm$ 1.7} & 66.0 $\pm$ 3.6 \\
Walker2d & M  & 74.0 $\pm$ 1.4          & 75.4 $\pm$ 1.1 & 73.0 $\pm$ 1.9 & \textbf{78.3 $\pm$ 1.5} & 74.1 $\pm$ 1.2 \\
\bottomrule
\end{tabular}
}
\end{table}

%% file: sections/05_results.tex
\section{Experiments}
\label{sec:experiments}

We evaluate Re:Frame on standard continuous-control benchmarks from MuJoCo~\citep{todorov2012mujoco}, a widely used physics simulator for RL research. For our evaluation, we use the D4RL offline datasets~\citep{fu2020d4rl}, which provide a standardized suite of trajectories with varying levels of quality. We focus on two commonly studied locomotion domains, \texttt{Hopper} and \texttt{Walker2d}, and consider the \texttt{Medium} (M) and \texttt{Medium-Replay} (MR) splits. Performance is reported in terms of D4RL normalized scores (mean~$\pm$~std across three independent runs). Model and training hyperparameters are available in~\autoref{appendix},~\autoref{tab:rate_hyperparams}.

In the D4RL setting, the \texttt{Medium} dataset consists of trajectories collected by a policy that achieves roughly one-third of the score of a fully expert policy, providing data of moderate but non-optimal quality. The \texttt{Medium-Replay} dataset, in contrast, corresponds to the entire replay buffer of an agent trained to that same medium level, which results in a broader and more heterogeneous distribution of behaviors that can be significantly harder to learn from. For reference, the \texttt{Expert} (E) datasets contain trajectories generated by a fully trained policy that achieves near-optimal performance, and therefore represent the upper bound of demonstration quality.  

As our baseline, we use the DT, a widely adopted offline RL method that frames RL as a sequence modeling problem. To assess the contribution of expert memory, Re:Frame augments DT with an AMB that contains only 60 expert trajectories. Notably, this represents just $0.1\%$ of a 6000-trajectory dataset, highlighting that Re:Frame is designed to operate effectively in the regime where expert demonstrations are extremely scarce but can still be leveraged through associative retrieval.

\input{tables/amb_size_ablations}
Re:Frame uses an expert AMB during both training and evaluation while the policy itself is trained on non-expert data (M or MR). In this regime, the agent learns from large quantities of suboptimal trajectories while being able to query a tiny pool of expert experiences via associative retrieval. To better understand the role of expert memory, we also conduct two ablations: (i) replacing the training AMB with dataset-derived trajectories but swapping in expert memory only at evaluation, and (ii) using a dataset-matched AMB at both training and evaluation.

The results in~\autoref{tab:main-results} reveal several consistent patterns.
First, using an expert AMB throughout training and evaluation yields clear gains in three of four benchmarks: +10.7 points on \texttt{Hopper-M}, +3.2 on \texttt{Walker2d-MR}, and +4.3 on \texttt{Walker2d-M}. This supports our hypothesis that even a very small set of high-quality expert trajectories can substantially improve learning from predominantly suboptimal data. Second, \texttt{Hopper-MR} remains challenging: DT achieves the best mean return, while Re:Frame reduces variance but does not match DT’s peak performance, likely due to the broader behavior-policy distribution in MR datasets. Third, simply introducing expert memory only at evaluation (swap setting) is ineffective, as agents must learn to exploit expert associations during training. Finally, when memory is derived from the same dataset as training (dataset AMB), performance closely follows DT, showing that the quality of stored experiences, rather than their quantity, drives improvement. In summary,~\autoref{tab:main-results} demonstrates that a lightweight, expert-driven associative memory (only 0.1\% of the dataset) can improve offline RL performance without modifying the backbone policy.

\subsection{AMB Size Ablations}
To further investigate Re:Frame’s dependence on the number of expert trajectories stored in memory, we varied the AMB size in the \textbf{expert train+eval} setting, progressively reducing it from 60 to 45 and finally to 30 trajectories. Results are reported in~\autoref{tab:amb-size-ablation}. As expected, overall performance degrades as the buffer shrinks, but the rate of decline is highly task-dependent.  

For \texttt{Hopper}, the agent shows strong sensitivity to the size of the AMB. Reducing the number of expert trajectories from 60 to 45 leads to a substantial performance drop, and with only 30 trajectories the agent nearly collapses on the MR split (69.7 $\rightarrow$ 40.0 $\rightarrow$ 3.0). This sharp decline indicates that the Hopper domain requires broad expert coverage in latent space for associative retrieval to succeed, and that the memory becomes too sparse when the number of examples is small.

In contrast, \texttt{Walker2d} demonstrates greater robustness. On the M split, performance remains essentially unchanged across all three buffer sizes (78.3 $\rightarrow$ 78.2 $\rightarrow$ 77.8), suggesting that even a limited number of expert trajectories is sufficient for effective retrieval. On the MR split, performance does decline as the AMB shrinks (69.8 $\rightarrow$ 67.6 $\rightarrow$ 45.4), but the degradation is more gradual compared to Hopper. This indicates that Walker2d’s expert trajectories are more homogeneous and thus require fewer examples to maintain adequate coverage in latent space.  

Taken together, these findings suggest that Re:Frame’s effectiveness is influenced not only by the presence of expert data, but also by how well the stored trajectories span the relevant latent space of expert behaviors. When the AMB becomes too small, associative retrieval fails to provide consistent guidance, particularly in tasks with more diverse or noisy behavior distributions such as Hopper-MR. In contrast, tasks with more structured or coherent expert behaviors, like Walker2d, can tolerate smaller memory sizes without significant loss in performance.

\subsection{DT Finetune}
To further examine whether Re:Frame’s improvements are simply the result of additional access to expert trajectories, we conducted a control experiment where the DT baseline was fine-tuned directly on the same 60 expert trajectories that were used to construct the AMB. This setup isolates the effect of expert data exposure without associative retrieval.  

As shown in~\autoref{tab:main-results}, fine-tuning DT provides only limited and inconsistent gains. In particular, it yields a small improvement on \texttt{Hopper-MR} (+2.1), but on other datasets the benefits are minimal or even negative. For example, on \texttt{Hopper-M} fine-tuning improves performance by only +0.7, while on \texttt{Walker2d-MR} performance actually drops below the original DT baseline. These results indicate that directly training on a handful of expert trajectories is insufficient for stable improvements, as the model easily overfits to the small set of demonstrations and fails to generalize across the broader dataset.  

In contrast, Re:Frame leverages the exact same 60 expert trajectories much more effectively. By storing them in the AMB and retrieving relevant latent representations during training and evaluation, the agent is able to repeatedly access and integrate expert information in a context-dependent manner. This continual and associative use of memory transforms scarce demonstrations into a reliable source of guidance.  

This highlights that the benefit of Re:Frame does not come from sheer exposure to expert data, but rather from the associative retrieval mechanism, which allows expert experience to be integrated throughout the entire learning process. In this way, Re:Frame turns a tiny expert subset into a persistent signal that improves both training stability and final performance.

%% file: tables/amb_size_ablations.tex
\begin{table}[t]
\centering
\caption{\textbf{Re:Frame (expert train+eval): effect of AMB size.} 
D4RL normalized scores (mean $\pm$ std) when varying the number of expert trajectories in the Associative Memory Buffer.}
\vspace{5pt}
\label{tab:amb-size-ablation}
\small
\begin{tabular}{ccccc}
\toprule
\shortstack[l]{AMB size} & \shortstack[c]{Hopper MR} & \shortstack[c]{Hopper M} & \shortstack[c]{Walker2d MR} & \shortstack[c]{Walker2d M} \\
\midrule
60 & \textbf{69.7 $\pm$ 1.7} & \textbf{78.3 $\pm$ 1.6} & \textbf{69.8 $\pm$ 1.7} & \textbf{78.3 $\pm$ 1.5} \\
45 & 40.0 $\pm$ 1.1 & 50.8 $\pm$ 0.3 & 67.6 $\pm$ 2.1 & 78.2 $\pm$ 0.3 \\
30 & 3.0 $\pm$ 0.1 & 42.8 $\pm$ 0.3 & 45.4 $\pm$ 1.1 & 77.8 $\pm$ 2.3 \\
\bottomrule
\end{tabular}
\vspace{-10pt}
\end{table}

%% file: sections/07_limitations.tex
\section{Limitations and Future Work}
\label{sec:limitations}

Our experiments show that Re:Frame can improve offline RL with very few expert demonstrations, but some limitations remain. Validation is currently limited to DT; applying Re:Frame to other offline RL methods such as IQL or TD3+BC would better demonstrate generality. 
Evaluation is restricted to locomotion benchmarks with medium- and replay-quality datasets. Extending to domains with sparse rewards, safety constraints, or abundant expert data would broaden applicability and clarify whether benefits persist beyond the low-expert regime. 
Finally, retrieval relies on nearest-neighbor similarity in the latent AMB space. Alternative strategies, scalable indexing~\citep{johnson2019billion}, or adaptive memory updates~\citep{graves2016hybrid} could enhance efficiency and robustness. Beyond offline RL, applying Re:Frame in online or hybrid settings is a promising direction, where small expert memories may guide exploration and speed up learning.

%% file: sections/08_conclusion.tex
\section{Conclusion}
\label{sec:conclusion}

In this work, we introduced \textbf{Re:Frame}, a lightweight and architecture-agnostic associative memory framework for offline RL. By augmenting a standard policy with a compact memory buffer containing only about 0.1\% expert trajectories, Re:Frame substantially improves decision-making when training mainly on suboptimal data. Our experiments on D4RL locomotion benchmarks show consistent gains over Decision Transformer in the limited-expert regime, with improvements of up to +10.7 normalized points. Overall, Re:Frame highlights the importance of compact expert memory in bridging the gap between scarce demonstrations and abundant suboptimal data. 
Because it requires no changes to the backbone model and relies on efficient retrieval, Re:Frame is simple to adopt and computationally lightweight. 
We hope these findings inspire exploration of expert memory across diverse domains, including online and hybrid RL.

%% file: sections/appendix/appendix.tex
\section{Supplementary Materials}
\label{appendix}

\begin{table}[H]
\begin{center}
\caption{Re:Frame hyperparameters.}
\label{tab:rate_hyperparams}
\vspace{5pt}
\begin{adjustbox}{width=0.5\textwidth}
\begin{tabular}{lccccccc}
\toprule
\textbf{Hyperparameter} &
\textbf{MuJoCo} \\
\midrule
\multicolumn{2}{l}{\textit{Transformer architecture}} \\
\midrule
Number of layers             & 3 \\
Number of attention heads    & 1 \\
Embedding dimension          & 128 \\
Context length $K$           & 60 \\
\midrule
\multicolumn{2}{l}{\textit{Regularization}} \\
\midrule
Hidden dropout               & 0.2 \\
Attention dropout            & 0.05 \\
Weight decay                 & 0.1 \\
\midrule
\multicolumn{2}{l}{\textit{Training configuration}} \\
\midrule
Training steps               & 500000 \\
Batch size                   & 50 \\
Optimizer                    & AdamW \\
Learning rate                & 6e-5 \\
Grad norm clip               & 1.0 \\
Linear warmup                & True \\
Warmup steps                 & 100 \\
Cosine decay                 & False \\
$(\beta_1, \beta_2)$         & (0.9, 0.95) \\
Loss function                & MSE \\
Max episode length           & 1000 \\
\bottomrule
\end{tabular}
\end{adjustbox}
\end{center}
\end{table}